\documentclass{article}

\usepackage{xcolor}
\usepackage[preprint]{corl_2026} % Pre-print version (shows authors, no CoRL footnote).

\usepackage{iftex}
\ifPDFTeX\else
  \usepackage{fontspec}
\fi

\usepackage{booktabs}
\usepackage{graphicx}
\usepackage{placeins}
\usepackage{float}
\usepackage{amsmath}

\newcommand{\pizero}{$\pi_0$}
\newcommand{\pihalf}{$\pi_{0.5}$}

\title{How Many Demonstrations Does a VLA Need?\\On the Sample Efficiency of VLA Finetuning}

\author{
  Rahul Chand \\
  Stanford University \\
  \texttt{rahulsc@cs.stanford.edu}
  \And
  Miguel Villa Floran \\
  California Polytechnic State University \\
  \texttt{mavillaf@calpoly.edu}
}

\begin{document}
\maketitle

%===============================================================================

\begin{abstract}
Pretrained vision language action (VLA) models show strong zero-shot performance but they are rarely deployed directly and are often first post-trained on demonstrations from the target setting. Yet VLAs are typically compared by zero-shot or full-data fine-tuned success which reveals little about the data and compute required for downstream adaptation. We study adaptation efficiency for six VLAs (\pizero, \pihalf, and GR00T N1-N1.7) on four LIBERO suites, varying both demonstrations and training steps. This reveals differences hidden by full-data fine-tuning: models which are within two points on the full LIBERO-Spatial dataset differ by up to 3$\times$ in success with 50 demonstrations. \pihalf{} needs 2-5$\times$ fewer demonstrations than \pizero{} and is the most sample-efficient model at every budget, with the largest gap on the long-horizon suite. Within GR00T, N1.5 adapts more efficiently than N1.6 and N1.7 on three of four suites, despite their stronger zero-shot performance. We also study one pretraining factor behind these differences by freezing the VLM during finetuning: this severely degrades the $\pi$ models whose VLM is jointly trained with the action expert during pretraining, but barely affects GR00T models.
\end{abstract}

%===============================================================================

\section{Introduction}
\label{sec:intro}

State-of-the-art robot manipulation policies are typically obtained in two stages. A generalist policy is first pretrained on large and diverse robot and web data, and is then fine-tuned on demonstrations collected in the target deployment setting. A pretrained policy's value depends on two things: how well it works out of the box, and how cheaply it can be adapted to a new setting, in both demonstrations and compute. Despite this, VLAs are usually compared through zero-shot evaluations or through success after fine-tuning on a full benchmark dataset~\citep{black2025pi0,physicalintelligence2025pi05,nvidia2025grootn1,bjorck2025grootn15}. Neither number measures the cost of adaptation. Yet many practical deployments have only a limited number of demonstrations, which makes adaptability under a demonstration budget an important measure of a pretrained policy.

Full-data fine-tuning hides large differences between models: on LIBERO-Spatial, five of the six models we study reach 97–99\% with the full dataset, but only 31–94\% with 50 demonstrations. Models with stronger reported zero-shot performance, trained on more data with more parameters can match their predecessors at full data and still require up to twice the data or compute to get there.

We present a systematic study of VLA adaptation efficiency under behavior cloning (which is usually the standard way pretrained VLAs are fine-tuned on new tasks). We compare two families. \pizero{}~\citep{black2025pi0} and \pihalf{}~\citep{physicalintelligence2025pi05} share the same PaliGemma VLM backbone and a flow-matching action expert, and differ primarily in their pretraining recipe. The GR00T N1, N1.5, N1.6, and N1.7 models~\citep{nvidia2025grootn1,bjorck2025grootn15,nvidia2025grootn16,nvidia2026grootn17} are successive iterations, consisting of a VLM paired with a diffusion-transformer action head. Each reports to improve on its predecessor's generalist performance. Within each family, the pretrained models act as a natural experiment over pretraining recipes, which would otherwise require prohibitively expensive pretraining ablations.

Our contributions are:
\begin{itemize}
    \item \textbf{A large-scale study of VLA sample efficiency.} We fine-tune six VLAs on four LIBERO suites of increasing difficulty and measure adaptation along two axes: the number of demonstrations needed to reach a given success rate, and the number of optimization steps needed with the full dataset. The study comprises 120 fine-tuning runs and nearly 1400 checkpoint evaluations, more than 550k simulated episodes in total (Sections~\ref{sec:setup}-\ref{sec:results-efficiency}).
    \item \textbf{Evidence that zero-shot and full-data comparisons do not capture adaptation efficiency.} \pihalf{} is 2--5$\times$ more demonstration-efficient than \pizero{} and outperforms every GR00T model at every budget, with the gap growing on the long-horizon suite. Within GR00T, N1.5 outperforms the stronger zero-shot models N1.6 and N1.7 at every tested budget on Spatial, Object, and Goal.
    \item \textbf{A downstream probe of one pretraining choice.} Ablating pretraining recipes is prohibitively expensive, so we study one choice whose effect is visible during adaptation: whether the VLM is frozen or trained jointly with the action head during pretraining. Freezing the VLM during fine-tuning is catastrophic for the $\pi$ family, which trains its VLM jointly, but nearly free for GR00T models, which keep it frozen (Section~\ref{sec:results-freeze}).
\end{itemize}

%===============================================================================

\section{Related Work}
\label{sec:related}

\paragraph{Pretrained VLAs.} Generalist VLAs couple a pretrained VLM with an action decoder trained on large robot datasets. \pizero{}~\citep{black2025pi0} attaches a flow-matching action expert to PaliGemma. \pihalf{}~\citep{physicalintelligence2025pi05} keeps this architecture and changes the pretraining recipe, co-training on heterogeneous robot, web, and high-level subtask data. GR00T N1~\citep{nvidia2025grootn1} pairs a VLM with a diffusion-transformer action head, and its successors~\citep{bjorck2025grootn15,nvidia2025grootn16,nvidia2026grootn17} revise the backbone, the action head, and the training data. Other pretraining choices include action tokenization~\citep{pertsch2025fast} and implicit world modeling~\citep{zheng2025flare}.

\paragraph{Adapting VLAs.} A growing body of work improves how VLAs are adapted, through better fine-tuning recipes~\citep{kim2025oft}, parameter-efficient tuning~\citep{ferchau2026lora}, object-centric or future-conditioned adaptation for few-shot settings~\citep{li2025controlvla,nguyen2026foca}, one-shot skill adaptation~\citep{zhao2025mosvla}, and regularizers that preserve pretrained representations during fine-tuning~\citep{huang2026maps,kachaev2025dontblind}. These methods typically take the pretrained model as given. We study a complementary question: under behavior cloning which is usually the default adaptation procedure how strongly does the choice of pretrained model determine the data and compute needed to learn on new tasks?

\paragraph{Data efficiency and pretraining for adaptation.} Data quantity and quality are known to dominate imitation learning outcomes~\citep{mandlekar2022robomimic,lin2025datascaling}, and the choice of VLM backbone affects VLA performance~\citep{zhang2026vlm4vla}. In the low-data regime, retrieval from large prior datasets can augment a small set of target demonstrations~\citep{xie2025iwr}. In language modeling, better pretraining metrics do not always yield a model that fine-tunes better~\citep{springer2025overtrained}. We provide a systematic measurement of this kind for robot policies, across two model families, four task suites, and both the data and compute axes of adaptation.

%===============================================================================

\section{Experimental Setup}
\label{sec:setup}
\paragraph{Models.} We study \pizero{} and \pihalf{} from the openpi release and GR00T N1, N1.5, N1.6, and N1.7, starting each run from the publicly released base checkpoint. Our default adaptation fine-tunes all parameters. In the frozen-VLM ablation (Section~\ref{sec:results-freeze}), we freeze the vision encoder and language model and train only the remaining parameters (the action head) about 578M of 3.50B parameters for \pizero{} and 693M of 3.62B for \pihalf, and 1.07B, 1.42B, and 1.62B for GR00T N1.5, N1.6, and N1.7.

\begin{figure}[t]
\centering
\includegraphics[width=\textwidth]{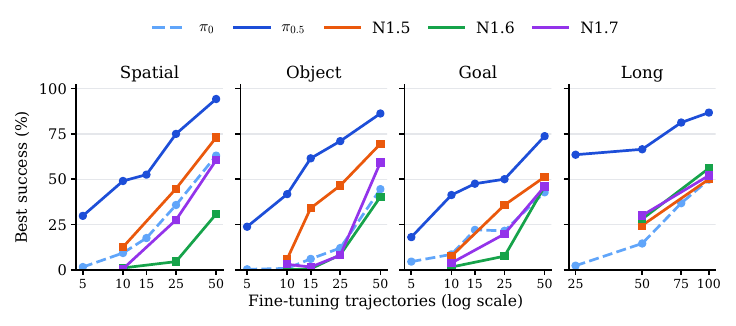}
\caption{Best checkpoint success versus number of fine-tuning demonstrations (whole suite, log scale). A model lying above another at a smaller budget is more sample-efficient: e.g., on Spatial, \pihalf{} with 25 demonstrations (75.0\%) outperforms \pizero{} with 50 (63.0\%). Within GR00T, N1.5 lies above N1.6 and N1.7 on Spatial, Object, and Goal.}
\label{fig:budget}
\end{figure}

\paragraph{Benchmarks.} We use the four LIBERO suites~\citep{liu2023libero}, ordered from easier to harder: Spatial, Object, and Goal each vary a single factor (object layout, object identity, and task goal, respectively) across ten short-horizon tasks, while Long (LIBERO-10) consists of ten long-horizon tasks that chain multiple subtasks. The training sets contain 432, 454, 428, and 379 demonstrations, respectively.

\paragraph{Demonstration budgets.} For the sample-efficiency sweeps, we fine-tune each model on the same $B$ demonstrations from the suite's training set with smaller budgets nested within larger ones. The $\pi$ runs use seed 42 and the GR00T runs seed 43. $B$ counts demonstrations for the whole ten-task suite and not per task. For the $\pi$ models we use $B\in\{5,10,15,25,50\}$ on Spatial, Object, and Goal, and $B\in\{25,50,75,100\}$ on Long, where five demonstrations per suite are not enough to make progress. Because of compute limits, GR00T N1.5, N1.6, and N1.7 are run with $B\in\{10,15,25,50\}$ on Object, $B\in\{10,25,50\}$ on Spatial and Goal, and $B\in\{50,100\}$ on Long. the remaining budgets are in progress and will be included in a revised version of this paper. In the \emph{full-data} setting we fine-tune on the complete training set and track success over training steps. Full-data fine-tuning is thus the upper end of our budget axis.

\paragraph{Optimization.} All runs use AdamW with an effective batch size of 48, so a training step processes the same number of samples for every model and steps are directly comparable as a measure of adaptation compute. The $\pi$ models use a learning rate of $5\times10^{-5}$. GR00T uses a learning rate and weight decay of $10^{-5}$ with 5\% warmup, cosine decay, and gradient clipping at 1. Full-data and budget runs share these settings. The budget sweeps train for 6-8 epochs ($\pi$) and 6-7 epochs (GR00T). Full-data runs are trained until performance approximately saturates, resulting in run lengths from 1.6k steps (\pihalf{} on Spatial, about 1.5 epochs) to 16.9k steps (\pizero{} on Long, 8 epochs).

\paragraph{Evaluation and metrics.} Each checkpoint is evaluated on 400 episodes (40 per sub-task), and we report the fraction of successful episodes. We summarize each run with three quantities:
\begin{itemize}
    \item \textbf{Best success:} the highest success rate across the evaluated checkpoints of a run.
    \item \textbf{Data-efficiency multiplier:} for a reference model at budget $B$, let $B'$ be the smallest tested budget at which a compared model reaches at least the reference's best success. We report $B/B'$. Since $B'$ is restricted to tested budgets  the multiplier is conservative and becomes a lower bound when $B'$ is the smallest tested budget.
    \item \textbf{Steps to $X\%$:} in the full-data setting, the first training step at which success is at least $X\%$ for two consecutive evaluations, for $X\in\{70,80,90\}$.
\end{itemize}

%===============================================================================

\section{How Sample-Efficient Are Pretrained VLAs?}
\label{sec:results-efficiency}

\begin{figure}[b]
\centering
\includegraphics[width=\textwidth]{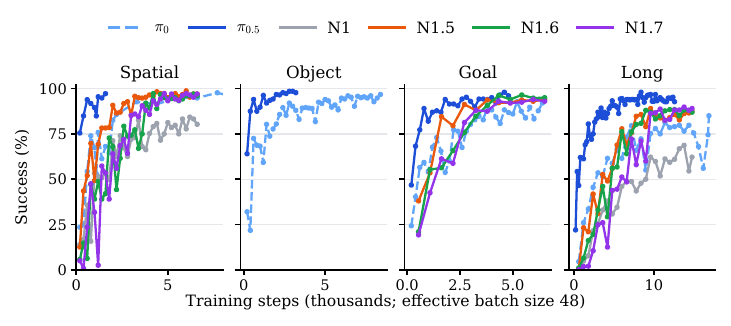}
\caption{Full-data fine-tuning of all six models (400 evaluation episodes per checkpoint). Most models converge to similar success on the short-horizon suites, but \pihalf{} gets there 2--5$\times$ faster than any other model. On Long it plateaus about 8 points above GR00T and 13 above \pizero. GR00T full-data runs on Object are not yet available.}
\label{fig:full}
\end{figure}

\subsection{\pizero{} versus \pihalf}
\label{sec:pi}

\paragraph{\pihalf{} needs 2--5$\times$ fewer demonstrations.} Figure~\ref{fig:budget} and Table~\ref{tab:multiplier} compare the two models across budgets. On every suite and at every budget, \pihalf{} reaches a higher best success than \pizero. Comparing tested budgets directly, \pihalf{} with 25 demonstrations outperforms \pizero{} with 50 on Spatial (75.0\% vs.\ 63.0\%), and \pihalf{} with 15 outperforms \pizero{} with 50 on Object (61.5\% vs.\ 44.5\%) and Goal (47.5\% vs.\ 42.8\%). At the low end \pihalf{} with only 5 demonstrations outperforms \pizero{} with 25 on Object (23.8\% vs.\ 12.0\%).

\paragraph{The gap widens on long-horizon tasks.} On LIBERO-Long \pihalf{} with the smallest budget we tested (25 demonstrations 63.5\%) already outperforms \pizero{} with 100 demonstrations (49.8\%), a multiplier of at least 4$\times$. Because of compute limits we did not run \pihalf{} below 25 demonstrations on Long so it may be even more sample-efficient there.

\begin{table}[t]
\centering
\caption{\textbf{How many demonstrations does \pihalf{} need to match each model?} Each cell reads ``best success of the column model with $B$ demonstrations $\rightarrow$ smallest tested budget $B'$ at which \pihalf{} reaches it ($B/B'$).'' Pairwise multipliers, including within GR00T, are listed in Table~\ref{tab:multiplier-full}.}
\label{tab:multiplier}
\scriptsize\setlength{\tabcolsep}{3pt}
\begin{tabular}{lr cccc}
\toprule
Suite & $B$ & $\pi_0$ & N1.5 & N1.6 & N1.7 \\
\midrule
Spatial & 25 & 35.8 $\rightarrow$ 10 (2.5$\times$) & 44.8 $\rightarrow$ 10 (2.5$\times$) & 4.5 $\rightarrow$ 5 (5.0$\times$) & 27.5 $\rightarrow$ 5 (5.0$\times$) \\
Spatial & 50 & 63.0 $\rightarrow$ 25 (2.0$\times$) & 73.0 $\rightarrow$ 25 (2.0$\times$) & 30.8 $\rightarrow$ 10 (5.0$\times$) & 60.5 $\rightarrow$ 25 (2.0$\times$) \\
Object & 25 & 12.0 $\rightarrow$ 5 (5.0$\times$) & 46.5 $\rightarrow$ 15 (1.7$\times$) & 8.2 $\rightarrow$ 5 (5.0$\times$) & 8.0 $\rightarrow$ 5 (5.0$\times$) \\
Object & 50 & 44.5 $\rightarrow$ 15 (3.3$\times$) & 69.5 $\rightarrow$ 25 (2.0$\times$) & 40.2 $\rightarrow$ 10 (5.0$\times$) & 59.2 $\rightarrow$ 15 (3.3$\times$) \\
Goal & 25 & 21.5 $\rightarrow$ 10 (2.5$\times$) & 35.8 $\rightarrow$ 10 (2.5$\times$) & 7.5 $\rightarrow$ 5 (5.0$\times$) & 19.8 $\rightarrow$ 10 (2.5$\times$) \\
Goal & 50 & 42.8 $\rightarrow$ 15 (3.3$\times$) & 51.2 $\rightarrow$ 50 (1.0$\times$) & 46.0 $\rightarrow$ 15 (3.3$\times$) & 46.0 $\rightarrow$ 15 (3.3$\times$) \\
Long & 50 & 14.5 $\rightarrow$ 25 (2.0$\times$) & 24.5 $\rightarrow$ 25 (2.0$\times$) & 28.0 $\rightarrow$ 25 (2.0$\times$) & 30.0 $\rightarrow$ 25 (2.0$\times$) \\
Long & 100 & 49.8 $\rightarrow$ 25 (4.0$\times$) & 50.0 $\rightarrow$ 25 (4.0$\times$) & 56.0 $\rightarrow$ 25 (4.0$\times$) & 52.0 $\rightarrow$ 25 (4.0$\times$) \\
\bottomrule
\end{tabular}

\end{table}

\paragraph{Full-data fine-tuning.} The full training set which is the upper end of our budget axis shows the same ordering in compute (Figure~\ref{fig:full}, Appendix Table~\ref{tab:steps}). The two models reach similar final performance on Spatial (97.7\% vs.\ 97.2\%) and Object (96.8\% vs.\ 98.5\%), but \pihalf{} reaches 80\% success after 0.4k steps on Spatial and Object, versus 2.0k for \pizero{} (5$\times$), and reaches 90\% on Goal after 1.8k steps versus 6.4k (3.6$\times$). On Long the difference extends to the ceiling: \pihalf{} reaches 80\% after 2.6k steps and peaks at 98\%, whereas \pizero{} first reaches 80\% only at its final checkpoint (16.9k steps) and peaks at 85\%. A comparison of final full-data success alone would therefore miss \pihalf's advantage on Spatial and Object entirely.
\section{Stronger zero-shot models are not more sample efficient: GR00T family}
\label{sec:groot}

Each GR00T model after N1 is reported to improve on its predecessor's generalist, zero-shot performance~\citep{bjorck2025grootn15,nvidia2025grootn16,nvidia2026grootn17}, with more pretraining data and progressively larger action heads (1.07B, 1.42B, and 1.62B parameters for N1.5, N1.6, and N1.7). If zero-shot capability translated into adaptability, sample efficiency should also improve from N1.5 to N1.7 but it does not.

\paragraph{N1.5 is the most sample-efficient GR00T model on Spatial, Object, and Goal.} At every tested budget on these three suites, N1.5 outperforms N1.6 and N1.7 (Figure~\ref{fig:budget}. All values in Appendix Table~\ref{tab:budget}). The gap is largest at small budgets: with 25 demonstrations, N1.5 reaches 44.8\%, 46.5\%, and 35.8\% on Spatial, Object, and Goal, versus 4.5--27.5\% for N1.6 and N1.7. N1.5 with 25 demonstrations even outperforms N1.6 with 50 on Spatial (44.8\% vs.\ 30.8\%) and Object (46.5\% vs.\ 40.2\%). Long is the exception, where N1.6 and N1.7 are slightly ahead at both tested budgets.

\paragraph{Training on full data recovers the difference.} With the full training sets (Figure~\ref{fig:full}, Appendix Table~\ref{tab:steps}), all three models end within a few points of each other on every suite (97--99\% on Spatial, 94--97\% on Goal, 88--90\% on Long), but they get there at different speeds. On Spatial, N1.5 reaches 90\% after 2.6k steps, whereas N1.6 and N1.7 need 4.6k and 4.2k (1.6-1.8$\times$ more). On Goal, N1.5 reaches 70\% after 1.6k steps versus 2.7k for both newer models. Consistent with the budget sweeps, the newer models are competitive only on the long-horizon suite. With limited demonstrations on the short-horizon suites the older N1.5 is clearly the better starting point.

\subsection{Across families: \pihalf{} versus GR00T}
\label{sec:cross}

\begin{figure}[t]
\centering
\includegraphics[width=\textwidth]{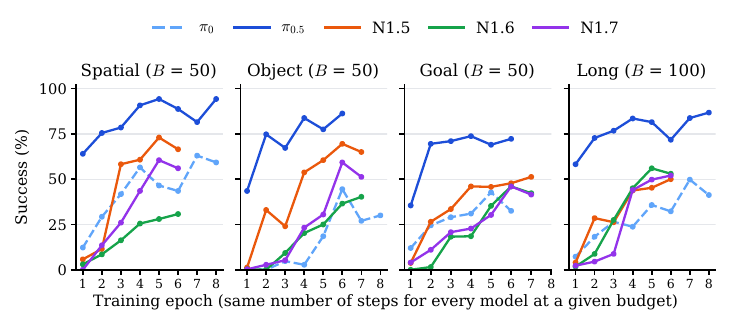}
\caption{Success over training with a fixed demonstration budget (50. 100 for Long). At a given budget, an epoch is the same number of optimization steps for every model. \pihalf{} leads from the first epoch on every suite. Among GR00T models, N1.5 learns fastest on Spatial, Object, and Goal. Curves for every budget are in Appendix~\ref{sec:curves}.}
\label{fig:budget-curves}
\end{figure}

Comparing families conflates many design choices, so we treat this comparison as indicative only. Both families share the evaluation protocol, but their learning-rate schedules differ (Section~\ref{sec:setup}). \pihalf{} outperforms every GR00T model at every tested budget on every suite (Figure~\ref{fig:budget}, Table~\ref{tab:multiplier}). The gap is largest on Long, where \pihalf{} with 25 demonstrations outperforms every GR00T model with 100. The advantage holds in compute at a fixed budget as well (Figure~\ref{fig:budget-curves}): after two epochs, \pihalf{} already exceeds the best success any GR00T model reaches in its whole run on every suite (e.g., 75.5\% vs.\ 73.0\% on Spatial and 72.8\% vs.\ 56.0\% on Long). With the full datasets (Figure~\ref{fig:full}, Appendix Table~\ref{tab:steps}), \pihalf{} reaches high success 2-4$\times$ faster than the fastest GR00T model on Spatial, Goal, and Long. \pizero{}, in contrast, trails the best GR00T model at every budget. On Object, N1.5 with 25 demonstrations already outperforms \pizero{} with 50 (46.5\% vs.\ 44.5\%). Since \pizero{} shares \pihalf's architecture, \pihalf's lead over GR00T comes from its pretraining rather than from the $\pi$ architecture.

%===============================================================================

\section{A Pretraining Choice Seen Through Adaptation: Freezing the VLM}
\label{sec:understanding}
\label{sec:results-freeze}

The results above show that models with similar architectures and similar full-data success can differ several-fold in how efficiently they adapt. This raises the question of which pretraining choices make a model adaptable. Answering it directly would require pretraining the same model under controlled variations of its recipe, which is prohibitively expensive. We therefore study one pretraining choice whose effect can be probed during downstream adaptation: whether the VLM is frozen during pretraining. The two families make opposite choices. GR00T keeps its VLM largely frozen during pretraining~\citep{nvidia2025grootn1,bjorck2025grootn15}, so its action head learns to work from fixed VLM features. The $\pi$ models instead train the VLM jointly with the action expert~\citep{black2025pi0,physicalintelligence2025pi05}. If this choice shapes adaptation, it should become visible when the VLM is also frozen during fine-tuning: a model pretrained with a frozen VLM should adapt well by updating its action head alone, whereas a model whose VLM and action expert co-adapted during pretraining should depend on updating the VLM. We test this by repeating our fine-tuning with the vision encoder and language model frozen, so that only the action head is trained (Section~\ref{sec:setup}).

\begin{figure}[t]
\centering
\includegraphics[width=\textwidth]{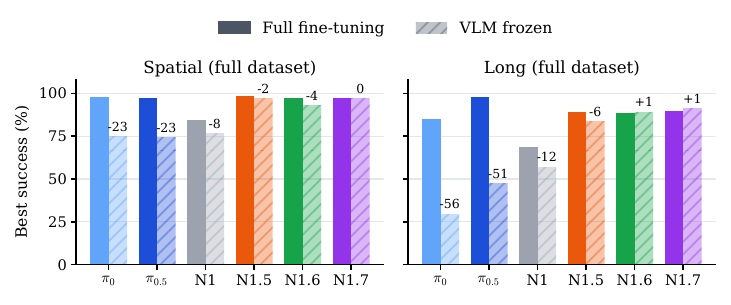}
\caption{Best success with full fine-tuning (solid) and with the VLM frozen (hatched) on the full Spatial and Long datasets. Numbers give the change in points. Freezing the VLM costs the $\pi$ models 23--56 points but changes GR00T N1.5--N1.7 by at most 6. Frozen runs train for 6 ($\pi$) and 8 (GR00T) epochs.}
\label{fig:freeze}
\end{figure}

\begin{figure}[t]
\centering
\includegraphics[width=\textwidth]{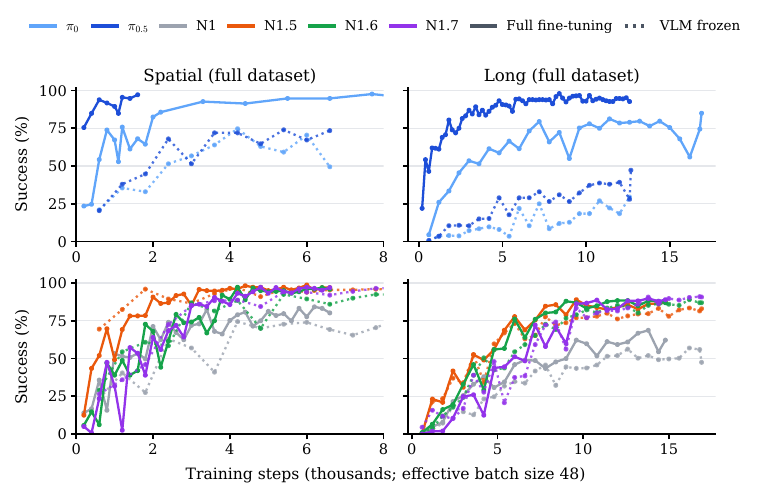}
\caption{Full-data learning curves with full fine-tuning (solid) and the VLM frozen (dotted). Freezing slows and caps the $\pi$ models, whereas the GR00T curves nearly coincide.}
\label{fig:frozen-curves}
\end{figure}

\paragraph{Sensitivity to freezing follows how the VLM was pretrained.} Freezing the VLM is costly for the $\pi$ models and nearly free for recent GR00T models (Figures~\ref{fig:freeze} and~\ref{fig:frozen-curves}). On Spatial, both \pizero{} and \pihalf{} lose 23 points (to 74-75\%). On Long, \pizero{} falls from 85\% to 30\% and \pihalf{} from 98\% to 47\%. In contrast, GR00T N1.5, N1.6, and N1.7 change by between $-6$ and $+1$ points on both suites (the small gains are within evaluation noise), and they learn at least as quickly as with full fine-tuning (e.g., 80\% on Spatial after 1.2k vs.\ 2.0k steps for N1.5. Appendix Table~\ref{tab:steps-frozen}). GR00T N1 is in between ($-8$ and $-12$). This is the pattern the pretraining choice predicts. A GR00T action head was pretrained to extract what it needs from fixed VLM features, and fine-tuning can do the same. In the $\pi$ models the VLM and action expert co-adapted during pretraining, so downstream adaptation also relies on changing the VLM, and freezing it removes a pathway the model learned to use. Because the two families also differ in data, architecture, and scale, this evidence is consistent with the effect of the pretraining choice but does not isolate it.

\paragraph{\pihalf's advantage needs an adaptable backbone on Spatial, but not on Long.} On Spatial, freezing the VLM removes the gap entirely (74.0\% vs.\ 74.8\% with the full dataset. Appendix Figure~\ref{fig:frozen-budget}). On Long, \pihalf{} still leads with the VLM frozen (47\% vs.\ 30\%), although both models fall well below their fully tuned performance. \textbf{A deployment that can only afford to update the action head is therefore well served by GR00T but not by the $\pi$ models.}

%===============================================================================

\section{Discussion and Conclusion}
\label{sec:discussion}

% \paragraph{Practical implications.} Whether the VLM was frozen during pretraining determines whether a model can be adapted by updating its action head alone. Restricting fine-tuning to the action head is nearly lossless for recent GR00T models but costs the $\pi$ models 23--56 success points.

\paragraph{Conclusion.} This paper studies how efficiently pretrained VLAs adapt to new tasks under behavior cloning, across six models and four LIBERO suites. Adaptation efficiency differs several-fold between models and is predicted neither by zero-shot capability nor by full-data success. \pihalf{} needs 2--5$\times$ fewer demonstrations than \pizero{}, and N1.5 adapts more efficiently than its successors N1.6 and N1.7. Freezing the VLM during fine-tuning suggests that how a model was pretrained matters for adaptation: models whose VLM was frozen during pretraining (GR00T) adapt well by updating only the action head while models whose VLM was trained jointly with the action head ($\pi$) also need the VLM updated. We therefore suggest reporting success as a function of fine-tuning demonstrations and of fine-tuning steps alongside zero-shot and full-data results.

\paragraph{Limitations.} All experiments are conducted in simulation on LIBERO. Experiments on additional simulation benchmarks and on real-robot tasks are in progress and will be included in a future version of this paper. Finally, released models differ in many ways at once, so our evidence on \emph{which} pretraining choices matter is correlational.

\bibliography{example}

%===============================================================================

\clearpage
\appendix

\section{Learning Curves}
\label{sec:curves}

\begin{figure}[h]
\centering
\includegraphics[width=\textwidth]{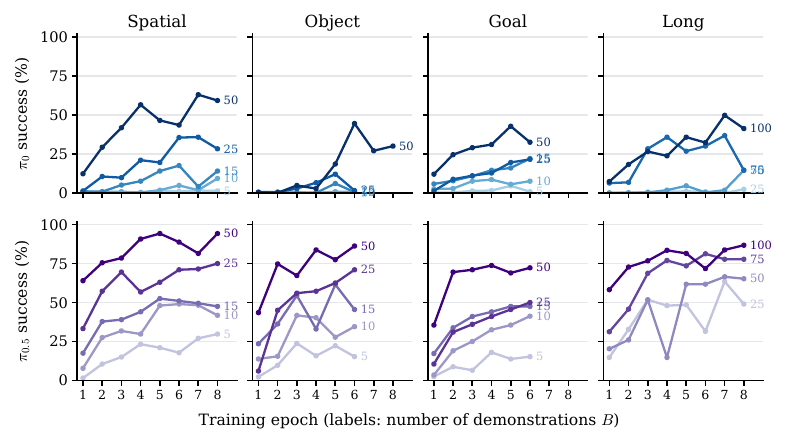}
\caption{\pizero{} (top) and \pihalf{} (bottom) budget sweeps with full fine-tuning, 400 evaluation episodes per checkpoint. Darker lines denote larger budgets.}
\label{fig:curves-pi}
\end{figure}

\begin{figure}[h]
\centering
\includegraphics[width=\textwidth]{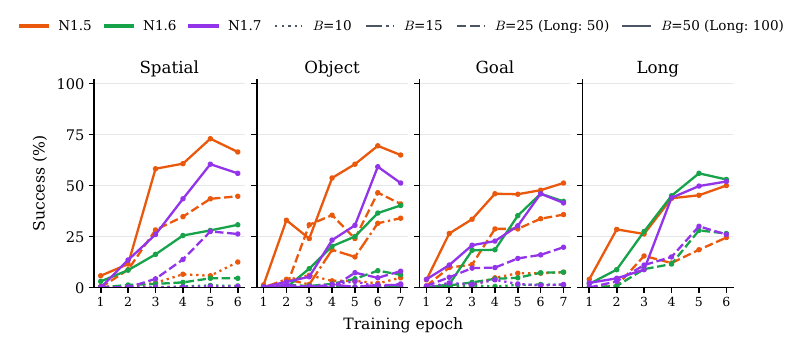}
\caption{GR00T budget sweeps with full fine-tuning, 400 evaluation episodes per checkpoint. Colors denote models and line styles budgets.}
\label{fig:curves-groot}
\end{figure}

Figures~\ref{fig:curves-pi}--\ref{fig:curves-groot} show success over training epochs for every budget, from which the best-checkpoint values in the main text are taken.

\section{Fine-Tuning with the VLM Frozen}
\label{sec:frozen-appendix}

\begin{figure}[h]
\centering
\includegraphics[width=\textwidth]{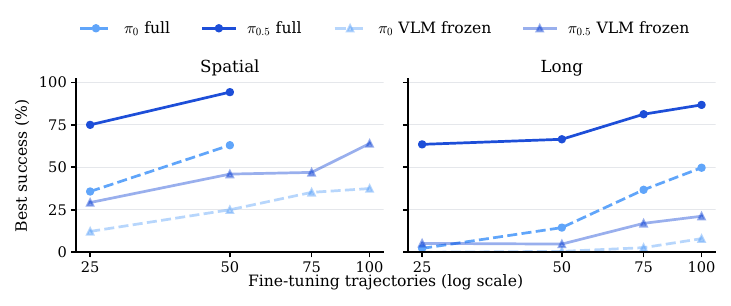}
\caption{Best success versus fine-tuning demonstrations for \pizero{} and \pihalf{} with full fine-tuning (opaque) and the VLM frozen (faded).}
\label{fig:frozen-budget}
\end{figure}

\begin{figure}[h]
\centering
\includegraphics[width=0.6\textwidth]{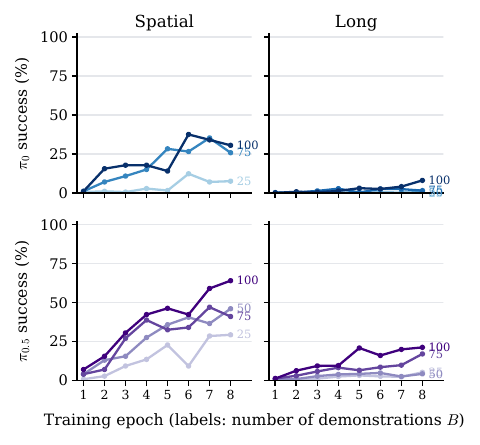}
\caption{\pizero{} (top) and \pihalf{} (bottom) budget sweeps with the VLM frozen. Only the best success (25.0\%) is available for the \pizero{} Spatial run with 50 demonstrations, so its curve is not shown.}
\label{fig:curves-frozen}
\end{figure}

\FloatBarrier
\section{Best Success at Every Budget}
\label{sec:budget-table}

\begin{table}[H]
\centering
\caption{Best checkpoint success (\%) at every fine-tuning budget $B$ (demonstrations per suite). N1.5--N1.7 denote GR00T models. ``--'' marks budgets not run.}
\label{tab:budget}
\small
\begin{tabular}{llrrrrr}
\toprule
\multicolumn{2}{l}{\textit{Spatial}} & $B{=}5$ & $B{=}10$ & $B{=}15$ & $B{=}25$ & $B{=}50$ \\
 & $\pi_0$ & 1.5 & 9.2 & 17.5 & 35.8 & 63.0 \\
 & $\pi_{0.5}$ & 29.8 & 49.0 & 52.5 & 75.0 & 94.2 \\
 & N1.5 & -- & 12.5 & -- & 44.8 & 73.0 \\
 & N1.6 & -- & 1.0 & -- & 4.5 & 30.8 \\
 & N1.7 & -- & 0.5 & -- & 27.5 & 60.5 \\
\midrule
\multicolumn{2}{l}{\textit{Object}} & $B{=}5$ & $B{=}10$ & $B{=}15$ & $B{=}25$ & $B{=}50$ \\
 & $\pi_0$ & 0.2 & 0.8 & 6.0 & 12.0 & 44.5 \\
 & $\pi_{0.5}$ & 23.8 & 41.8 & 61.5 & 71.0 & 86.2 \\
 & N1.5 & -- & 6.0 & 34.0 & 46.5 & 69.5 \\
 & N1.6 & -- & 0.0 & 0.5 & 8.2 & 40.2 \\
 & N1.7 & -- & 3.0 & 1.5 & 8.0 & 59.2 \\
\midrule
\multicolumn{2}{l}{\textit{Goal}} & $B{=}5$ & $B{=}10$ & $B{=}15$ & $B{=}25$ & $B{=}50$ \\
 & $\pi_0$ & 4.5 & 8.5 & 22.0 & 21.5 & 42.8 \\
 & $\pi_{0.5}$ & 18.0 & 41.2 & 47.5 & 50.0 & 73.8 \\
 & N1.5 & -- & 7.5 & -- & 35.8 & 51.2 \\
 & N1.6 & -- & 1.5 & -- & 7.5 & 46.0 \\
 & N1.7 & -- & 3.8 & -- & 19.8 & 46.0 \\
\midrule
\multicolumn{2}{l}{\textit{Long}} & $B{=}25$ & $B{=}50$ & $B{=}75$ & $B{=}100$ \\
 & $\pi_0$ & 2.2 & 14.5 & 36.8 & 49.8 \\
 & $\pi_{0.5}$ & 63.5 & 66.5 & 81.2 & 86.8 \\
 & N1.5 & -- & 24.5 & -- & 50.0 \\
 & N1.6 & -- & 28.0 & -- & 56.0 \\
 & N1.7 & -- & 30.0 & -- & 52.0 \\
\midrule
\multicolumn{2}{l}{\textit{Spatial (VLM frozen)}} & $B{=}25$ & $B{=}50$ & $B{=}75$ & $B{=}100$ \\
 & $\pi_0$ & 12.2 & -- & 35.2 & 37.5 \\
 & $\pi_{0.5}$ & 29.2 & 46.0 & 47.0 & 64.0 \\
\midrule
\multicolumn{2}{l}{\textit{Long (VLM frozen)}} & $B{=}25$ & $B{=}50$ & $B{=}75$ & $B{=}100$ \\
 & $\pi_0$ & 0.0 & 0.5 & 2.7 & 8.0 \\
 & $\pi_{0.5}$ & 5.2 & 4.8 & 17.0 & 21.2 \\
\bottomrule
\end{tabular}

\end{table}

\begin{table}[H]
\centering
\caption{All pairwise data-efficiency multipliers. For a reference model at budget $B$, $B'$ is the smallest tested budget at which the compared model reaches at least the reference's best success. Pairs where the compared model needs the full budget $B$ are omitted.}
\label{tab:multiplier-full}
\small
\begin{tabular}{llrr lrrr}
\toprule
& \multicolumn{3}{c}{Reference model} & \multicolumn{3}{c}{Compared model} & \\
\cmidrule(lr){2-4}\cmidrule(lr){5-7}
Suite & Model & $B$ & Best (\%) & Model & $B'$ & Best (\%) & $B/B'$ \\
\midrule
Spatial & $\pi_0$ & 25 & 35.8 & $\pi_{0.5}$ & 10 & 49.0 & 2.5$\times$ \\
Spatial & $\pi_0$ & 50 & 63.0 & $\pi_{0.5}$ & 25 & 75.0 & 2.0$\times$ \\
Spatial & N1.5 & 25 & 44.8 & $\pi_{0.5}$ & 10 & 49.0 & 2.5$\times$ \\
Spatial & N1.5 & 50 & 73.0 & $\pi_{0.5}$ & 25 & 75.0 & 2.0$\times$ \\
Spatial & N1.6 & 25 & 4.5 & $\pi_{0.5}$ & 5 & 29.8 & 5.0$\times$ \\
Spatial & N1.6 & 50 & 30.8 & $\pi_{0.5}$ & 10 & 49.0 & 5.0$\times$ \\
Spatial & N1.7 & 25 & 27.5 & $\pi_{0.5}$ & 5 & 29.8 & 5.0$\times$ \\
Spatial & N1.7 & 50 & 60.5 & $\pi_{0.5}$ & 25 & 75.0 & 2.0$\times$ \\
Spatial & N1.6 & 25 & 4.5 & N1.5 & 10 & 12.5 & 2.5$\times$ \\
Spatial & N1.6 & 50 & 30.8 & N1.5 & 25 & 44.8 & 2.0$\times$ \\
\midrule
Object & $\pi_0$ & 25 & 12.0 & $\pi_{0.5}$ & 5 & 23.8 & 5.0$\times$ \\
Object & $\pi_0$ & 50 & 44.5 & $\pi_{0.5}$ & 15 & 61.5 & 3.3$\times$ \\
Object & N1.5 & 25 & 46.5 & $\pi_{0.5}$ & 15 & 61.5 & 1.7$\times$ \\
Object & N1.5 & 50 & 69.5 & $\pi_{0.5}$ & 25 & 71.0 & 2.0$\times$ \\
Object & N1.6 & 25 & 8.2 & $\pi_{0.5}$ & 5 & 23.8 & 5.0$\times$ \\
Object & N1.6 & 50 & 40.2 & $\pi_{0.5}$ & 10 & 41.8 & 5.0$\times$ \\
Object & N1.7 & 25 & 8.0 & $\pi_{0.5}$ & 5 & 23.8 & 5.0$\times$ \\
Object & N1.7 & 50 & 59.2 & $\pi_{0.5}$ & 15 & 61.5 & 3.3$\times$ \\
Object & N1.6 & 25 & 8.2 & N1.5 & 15 & 34.0 & 1.7$\times$ \\
Object & N1.6 & 50 & 40.2 & N1.5 & 25 & 46.5 & 2.0$\times$ \\
Object & N1.7 & 25 & 8.0 & N1.5 & 15 & 34.0 & 1.7$\times$ \\
\midrule
Goal & $\pi_0$ & 25 & 21.5 & $\pi_{0.5}$ & 10 & 41.2 & 2.5$\times$ \\
Goal & $\pi_0$ & 50 & 42.8 & $\pi_{0.5}$ & 15 & 47.5 & 3.3$\times$ \\
Goal & N1.5 & 25 & 35.8 & $\pi_{0.5}$ & 10 & 41.2 & 2.5$\times$ \\
Goal & N1.6 & 25 & 7.5 & $\pi_{0.5}$ & 5 & 18.0 & 5.0$\times$ \\
Goal & N1.6 & 50 & 46.0 & $\pi_{0.5}$ & 15 & 47.5 & 3.3$\times$ \\
Goal & N1.7 & 25 & 19.8 & $\pi_{0.5}$ & 10 & 41.2 & 2.5$\times$ \\
Goal & N1.7 & 50 & 46.0 & $\pi_{0.5}$ & 15 & 47.5 & 3.3$\times$ \\
Goal & N1.6 & 25 & 7.5 & N1.5 & 10 & 7.5 & 2.5$\times$ \\
\midrule
Long & $\pi_0$ & 50 & 14.5 & $\pi_{0.5}$ & 25 & 63.5 & 2.0$\times$ \\
Long & $\pi_0$ & 75 & 36.8 & $\pi_{0.5}$ & 25 & 63.5 & 3.0$\times$ \\
Long & $\pi_0$ & 100 & 49.8 & $\pi_{0.5}$ & 25 & 63.5 & 4.0$\times$ \\
Long & N1.5 & 50 & 24.5 & $\pi_{0.5}$ & 25 & 63.5 & 2.0$\times$ \\
Long & N1.5 & 100 & 50.0 & $\pi_{0.5}$ & 25 & 63.5 & 4.0$\times$ \\
Long & N1.6 & 50 & 28.0 & $\pi_{0.5}$ & 25 & 63.5 & 2.0$\times$ \\
Long & N1.6 & 100 & 56.0 & $\pi_{0.5}$ & 25 & 63.5 & 4.0$\times$ \\
Long & N1.7 & 50 & 30.0 & $\pi_{0.5}$ & 25 & 63.5 & 2.0$\times$ \\
Long & N1.7 & 100 & 52.0 & $\pi_{0.5}$ & 25 & 63.5 & 4.0$\times$ \\
\bottomrule
\end{tabular}

\end{table}

\section{Full-Data Adaptation Speed}
\label{sec:steps-full}

\begin{table}[H]
\centering
\caption{Full-data fine-tuning: training steps (effective batch size 48) until success is at least 70/80/90\% at two consecutive evaluations, and best success. N1--N1.7 denote GR00T models. ``--'' means the threshold was not reached within the run. To save compute, some runs are evaluated less frequently (every 200--1,100 steps), which bounds the precision of these numbers.}
\label{tab:steps}
\small
\begin{tabular}{ll rrr r}
\toprule
Suite & Model & 70\% & 80\% & 90\% & Best (\%) \\
\midrule
Spatial & $\pi_0$ & 2.0k & 2.0k & 3.3k & 97.7 \\
 & $\pi_{0.5}$ & 0.2k & 0.4k & 0.6k & 97.2 \\
 & N1 & 3.0k & 6.2k & -- & 84.2 \\
 & N1.5 & 1.4k & 2.0k & 2.6k & 98.8 \\
 & N1.6 & 2.6k & 3.8k & 4.6k & 97.2 \\
 & N1.7 & 3.0k & 3.0k & 4.2k & 97.2 \\
\midrule
Object & $\pi_0$ & 1.4k & 2.0k & 2.8k & 96.8 \\
 & $\pi_{0.5}$ & 0.4k & 0.4k & 1.2k & 98.5 \\
\midrule
Goal & $\pi_0$ & 2.2k & 3.0k & 6.4k & 92.8 \\
 & $\pi_{0.5}$ & 0.6k & 0.8k & 1.8k & 98.0 \\
 & N1.5 & 1.6k & 2.7k & 3.8k & 94.0 \\
 & N1.6 & 2.7k & 3.3k & 3.8k & 96.5 \\
 & N1.7 & 2.7k & 2.7k & 4.3k & 94.0 \\
\midrule
Long & $\pi_0$ & 6.6k & 16.9k & -- & 85.0 \\
 & $\pi_{0.5}$ & 1.6k & 2.6k & 4.6k & 98.0 \\
 & N1 & -- & -- & -- & 68.7 \\
 & N1.5 & 7.2k & 7.8k & -- & 89.0 \\
 & N1.6 & 7.2k & 7.8k & -- & 88.4 \\
 & N1.7 & 9.6k & 9.6k & -- & 89.8 \\
\bottomrule
\end{tabular}

\end{table}

\begin{table}[H]
\centering
\caption{Frozen-VLM fine-tuning on the full datasets: training steps until success is at least 70/80/90\% at two consecutive evaluations, and best success. Same conventions as Table~\ref{tab:steps}.}
\label{tab:steps-frozen}
\small
\begin{tabular}{ll rrr r}
\toprule
Suite & Model & 70\% & 80\% & 90\% & Best (\%) \\
\midrule
Spatial & $\pi_0$ & -- & -- & -- & 74.8 \\
 & $\pi_{0.5}$ & 3.6k & -- & -- & 74.0 \\
 & N1 & 4.2k & -- & -- & 76.5 \\
 & N1.5 & 1.2k & 1.2k & 3.6k & 97.0 \\
 & N1.6 & 3.0k & 3.0k & 7.2k & 92.8 \\
 & N1.7 & 2.4k & 3.0k & 5.4k & 96.8 \\
\midrule
Long & $\pi_0$ & -- & -- & -- & 29.5 \\
 & $\pi_{0.5}$ & -- & -- & -- & 47.2 \\
 & N1 & -- & -- & -- & 57.0 \\
 & N1.5 & 7.8k & 15.6k & -- & 83.5 \\
 & N1.6 & 9.0k & 10.2k & -- & 89.0 \\
 & N1.7 & 7.8k & 10.8k & 16.2k & 91.0 \\
\bottomrule
\end{tabular}

\end{table}

\end{document}